\documentclass{article}

\usepackage{PRIMEarxiv}

\usepackage[utf8]{inputenc} 
\usepackage[T1]{fontenc}    
\usepackage{hyperref}       
\usepackage{url}            
\usepackage{booktabs}      
\usepackage{amsfonts}      
\usepackage{nicefrac}     
\usepackage{microtype}     
\usepackage{lipsum}
\usepackage{fancyhdr}      
\usepackage{graphicx}       
\usepackage{subcaption}
\graphicspath{{media/}}     

\usepackage[table,xcdraw]{xcolor}
\usepackage{multirow}

\title{Subgroup discovery of Parkinson's Disease by utilizing a multi-modal smart device system}

\author{
 Catharina Marie van Alen$^a$, Alexander Brenner$^a$, Tobias Warnecke$^b$, Julian Varghese$^a$ \\
  $^a$Institute of Medical Informatics, University of Münster, Münster, Germany\\
  $^b$Department of Neurology and Neurorehabilitation, Klinikum Osnabrück – Academic
teaching hospital of \\ the University of Münster, Osnabrück, Germany
}

\begin{document}
\maketitle

\begin{abstract}
In recent years, sensors from smart consumer devices have shown great diagnostic potential in movement disorders. In this context, data modalities such as electronic questionnaires, hand movement and voice captures have successfully captured biomarkers and allowed discrimination between Parkinson’s disease (PD) and healthy controls (HC) or differential diagnosis (DD). However, to the best of our knowledge, a comprehensive evaluation of assessments with a multi-modal smart device system has still been lacking. In a prospective study exploring PD, we used smartwatches and smartphones to collect multi-modal data from 504 participants, including PD patients, DD and HC. This study aims to assess the effect of multi-modal vs. single-modal data on PD vs. HC and PD vs. DD classification, as well as on PD group clustering for subgroup identification. We were able to show that by combining various modalities, classification accuracy improved and further PD clusters were discovered. 
\end{abstract}


\section{Introduction}
Parkinson's Disease (PD) is a neurodegenerative disorder with well-known symptoms such as slowed movement, rigidity, tremor and various non-motor symptoms (NMS). The appearance of these symptoms and the disease progress, however, highly differ from patient to patient and clinical documentation does not capture fine-grained objective phenotypical characteristics. Clinical documentation of motor symptoms, for instance, only describes three main PD subtypes: 1) Tremor-dominant PD, 2) Akineto-rigid PD, 3) Mixed/Equivalence type. Although there is no neuroprotective or regenerative treatment to date, early diagnosis and treatment is important in reducing burden and potential treatment costs \cite{postuma2019prodromal}. Thus, there is a need for early objective biomarkers.

Various systems have already demonstrated promising diagnostic potential when analyzing data modalities like electronic questionnaires, hand movement and voice captures \cite{lee2016validation, rusz2018smartphone, carignan2015measuring}. These studies were able to differentiate between PD and healthy subjects based on digital biomarkers, yielding an important step towards potential clinical adaptation. However, to the best of our knowledge, there is still a lack of comprehensive evaluation of combinations of these biomarkers in an interactive smart-device-based assessment setting. In particular, it is important to consider other similar movement disorders in the analysis as well to improve disease-specificity of the biomarkers.

To approach this problem, we have developed a Smart Device System (SDS) to analyze PD patients based on multi-modal data recording. In a compact assessment, the SDS was used to record self-completed electronic questionnaires and smartwatch-based sensor measures from a series of movement tasks. Given this system, we have recorded a total of 503 patient sessions in a prospective study from 2018 to 2021. Based on this study data, we have already found high diagnostic potential utilizing Machine Learning (ML) methods with movement and questionnaire data \cite{varghese2021sensor}. In a later stage of the study, further selected modalities were added to the assessments, these are speech recordings and a smartphone-based finger tapping task. 
In this work, we analyze the advances of the multi-modality of our system. We therefore 1) train ML models to discriminate PD from healthy controls and other movement disorders, and 2) perform cluster analysis within the PD group. Our research question is whether the usage of the multi-modal data compared to single-modal data increases information gain and thus 1) improves diagnostic accuracy when combined, and 2) lets us discover distinguishable PD subgroups that go beyond the aforementioned three clinically established main types.

\section{Study Data}
The study has been registered (ClinicalTrials.gov ID: NCT03638479) and approved by the ethical board of the University of Münster and the physician’s chamber of Westphalia-Lippe (Reference number: 2018-328-f-S).
Three participant groups have been recorded: 1) Parkinson’s disease (PD), including a broad range of different PD progress states according to Hoehn and Yahr \cite{bhidayasiri2012parkinson}, 2) differential diagnoses (DD) and 3) healthy controls (HC). Diagnoses were based on ICD-10 codes, confirmed by neurologists and reviewed by one senior movement disorder expert. 

Our analysis focuses on participants that completed an assessment that included all data modalities. From each participant we collected the following data:
\begin{enumerate}
    \item \textbf{Self-completed questionnaire:} The first part includes information about age, height, weight, kinship with PD, alcohol consume and medication. The second part consists of 30 yes/no items about NMS based on Chaudhuri et al. \cite{chaudhuri2006international}.
    \item \textbf{Smartwatch-recorded movement tasks:} 11 different movement tasks of 10 to 20 seconds length were performed with one smartwatch attached to each of the participants wrist respectively. Acceleration and rotation data were recorded synchronously.
    \item \textbf{Voice recording:} 3 types of speech tasks were recorded: (i) holding vowel tones ("a"/"i"/"o") per one breath, (ii) fast repetition of syllables "pah"/"tah"/"kah" and (iii) sentences reading.
    \item \textbf{Finger tapping:} Using 3 fingers, participants were asked to tap the smartphone screen repeatedly for 15 seconds as quick as possible.
\end{enumerate}

Details about the individual assessment steps are described in Varghese et al. \cite{varghese2019smart}. The sample size of all participants is summarized in \autoref{tab:0}.

\begin{table}[h!]
\caption{Participant sample}
\centering
\begin{tabular}{c|c|c}
\hline
Data modalities                                                                             & Disease class & Sample size \\ \hline
\multirow{3}{*}{\begin{tabular}[c]{@{}c@{}}Questionnaires, \\ movement\end{tabular}} & PD            & 279         \\
                                                                                            & DD            & 133         \\
                                                                                            & HC            & 90          \\ \hline
\multirow{3}{*}{\begin{tabular}[c]{@{}c@{}}Questionnaires,\\ movement, voice,\\ finger tapping\end{tabular}}                                                                   & PD            & 21          \\
                                                                                            & DD            & 27          \\
                                                                                            & HC            & 23          \\ \hline
\end{tabular}
\label{tab:0}
\end{table}

\subsection{Feature Extraction}  \label{sec:FeatureExtraction}
Given the assessment data, we performed a feature extraction procedure in order to prepare the data for ML. The following feature sets were generated for the respective modalities and used for classification:
\begin{enumerate}
    \item \textbf{Self-completed questionnaire:} All 30 NMS answers were used in binary format, other personal data was not considered.
    \item \textbf{Smartwatch-recorded movement tasks:} Two representative tasks were selected, "Relaxed" and "Lift and Hold". The recorded movements consist of time series for both smart-watches in three spatial axes for acceleration and rotation sensor measures. On these time series we computed frequency powers for 2 to 12 Hz in 1 Hz steps using Welch's power spectral density (PSD).
    \item \textbf{Voice recording:} We computed Jitter via autocorrelation on all vocal tasks, measuring the extent of variation of the voice range.
    \item \textbf{Finger tapping:} We divided the 15 seconds long record in three equal size segments and calculated the average speed and total count of display touches in every segment.
\end{enumerate}
In addition, we generated a subset for the cluster analysis to account for the small sample size of the PD group with all data modalities (see \autoref{tab:0}). 
Voice and finger tapping features were fully included in the subset. Questionnaire data were reduced to one feature by summing positively answered questions. Moreover, the movement features were reduced by only including the assessment "Relaxed" and summing the frequency powers from 2 to 12 Hz.
\section{Classification}
\begin{figure}[h!]
	\centering
		\includegraphics[width=0.6\textwidth]{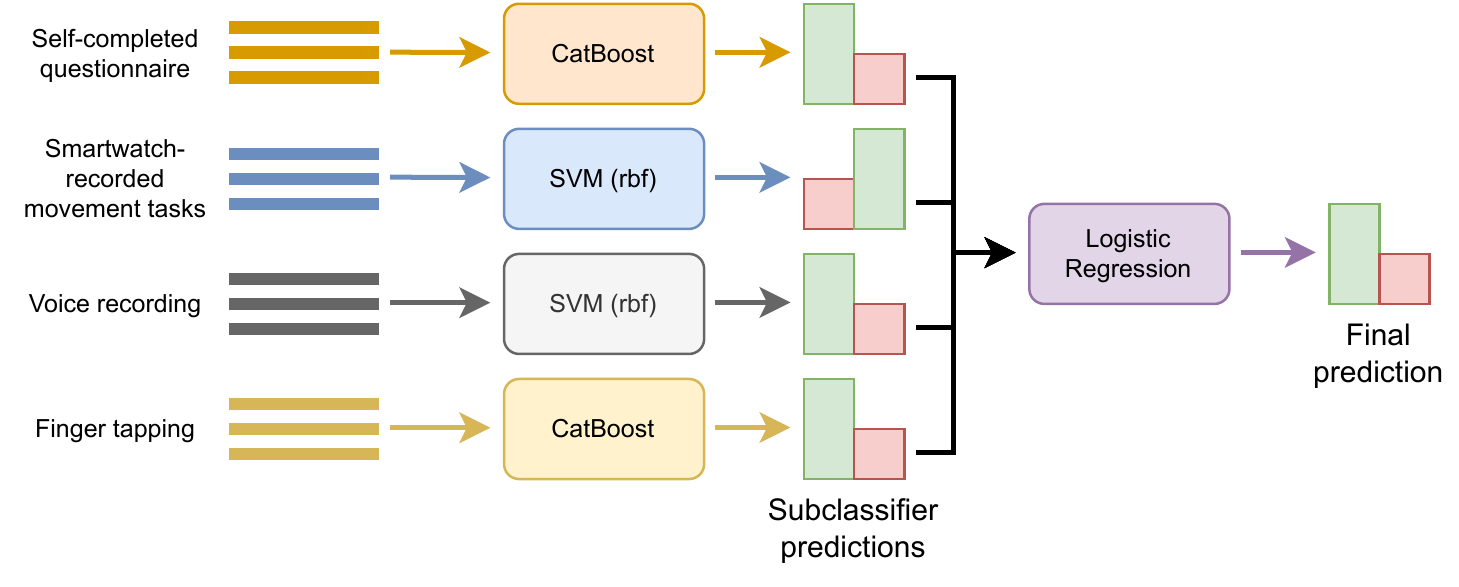}
	\caption{Stacking classifier for the classification of PD samples. For each data modality, a selected classifier is trained. The outputs of each subclassifier are forwarded to a logistic regression that performs the final classification.}
	\label{fig:architecture}
\end{figure}

Given the previously described features, we have trained and optimized ML classifiers for the individual data modalities. We used the scikit-learn implementation of the support-vector machine (SVM) \cite{sklearn_api} and CatBoost, a gradient boosting decision-tree-based model \cite{dorogush2018catboost}. To evaluate the potential information gain of combining features from different data modalities, we performed an adapted version of classifier stacking. In this version, a certain classifier is trained on each respective source of sensor data, e.g. the movement data is only fed to a movement classifier. In this way, we trained one classifier for each data modality respectively and thus can utilize the additional samples for smartwatch and questionnaire data in the training process. A simple linear model was trained on top of the individual outputs to consider all data modalities in the classification process and compute the final label for the input samples. \autoref{fig:architecture} summarizes the architecture and the utilized classifiers. To account for sample size differences, we used balanced class weighting in the training process and report results based on balanced classification accuracy. For evaluation, a 3 times randomly repeated 5-fold cross-validation was used. Two classification tasks were performed: PD vs. HC and PD vs. DD.

\subsection{Results}
Classification performance has been evaluated for the individual classifiers for each respective data modality and the combination of all features using the previously described stacking approach. \autoref{tab:classification} summarizes the averaged classification scores from the cross-validation.

\begin{table}[h!]
\caption{Evaluation of questionnaire, movement, voice and finger tapping data on the sample subset with all data records (44 samples for PD vs. HC, 48 samples for PD vs. DD). Performance is measured with balanced accuracy (STD). Best results are marked in bold.}
\centering
\begin{tabular}{l|l|l}
\hline
Task                            & PD vs. HC              & PD vs. DD              \\ \hline
Quest.                          & 0.843 (0.098)          & 0.667 (0.112)          \\ \hline
Mov.                            & 0.825 (0.112)          & 0.614 (0.102)         \\ \hline
Voice                           & 0.702 (0.154)          & 0.560 (0.198)          \\ \hline
Finger Tapping                  & 0.6383 (0.157)         & 0.570 (0.084)          \\ \hline
Quest. + Mov. + Voice + Finger Tapping & \textbf{0.918 (0.074)} & \textbf{0.722 (0.121)} \\ \hline
\end{tabular}
\label{tab:classification}
\end{table}

\section{Clustering}
We conducted hierarchical clustering within the PD group using the scikit-learn implementation of the agglomerative cluster algorithm \cite{sklearn_api}. To analyze information gain through multi-modality, we compared clustering results of a single data modality (movement features) with multiple data modalities (movement, voice, finger tapping and questionnaire features). The optimal number of clusters was determined from dendrograms by identifying the longest distance between joined clusters. For each cluster, we summarized the cluster composition by distinguishing between the clinically established PD types: The tremor-dominant type (T-type), the akineto-rigid type (AR-type) and the equivalence type (ART-type). These PD types were assigned to the participants by physicians in advance. Participants that could not be categorized to any of the types were documented as Unknown.   
\subsection{Results}
\begin{figure}[h]
\centering
\begin{subfigure}{0.5\textwidth}
  \centering
  \includegraphics[width=1\linewidth]{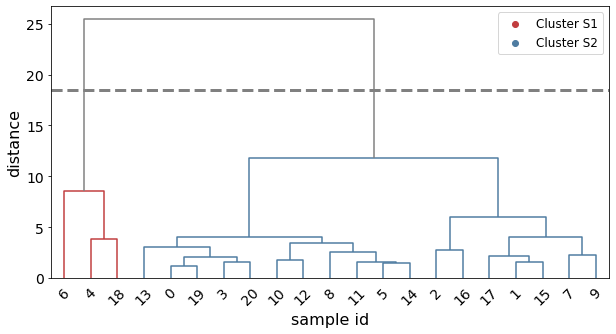}
  \caption{}
  \label{fig:sub1}
\end{subfigure}%
\begin{subfigure}{0.5\textwidth}
  \centering
  \includegraphics[width=1\linewidth]{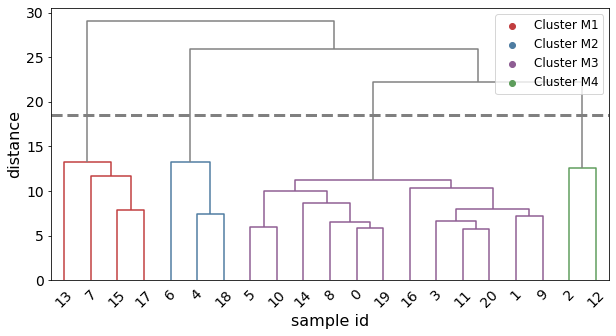}
  \caption{}
  \label{fig:sub2}
\end{subfigure}
\caption{Dendrograms of the hierarchical cluster analysis for (a) single-modal data (movement features) and (b) multi-modal data (movement, finger tapping, voice and questionnaire features). The gray horizontal line intersects the largest vertical distance between joined clusters.}
\label{fig:clusteringMov}
\end{figure}

\autoref{fig:clusteringMov} shows the dendrograms for (a) a single data modality (movement features) and (b) multiple data modalities (movement, finger tapping, voice and questionnaire features). In the single-modal analysis, clusters were labeled with the letter S (e.g. cluster S1), in the multi-modal analysis with the letter M (e.g. cluster M1). \autoref{tab:clusterMov} displays the corresponding cluster composition. 

\begin{table}[h]
\caption{The cluster composition by PD types corresponding to the cluster analysis in \autoref{fig:clusteringMov}. All values are given in percent and rounded to the second decimal place.}
\centering
\begin{tabular}{l  cc  cccc}
\toprule
 & \multicolumn{2}{c}{Movement} & \multicolumn{4}{c}{Movement, Finger Tapping, Voice, Questionnaire} \\
\cmidrule(lr){2-3} \cmidrule(lr){4-7}
Type & Cluster S1 & Cluster S2 & Cluster M1 & Cluster M2 & Cluster M3 & Cluster M4  \\
\midrule
T-Type &   33.33 & 0     & 0   & 33.33  & 0        & 0  \\
AR-Type &  0     & 44.44 & 25  & 0      & 50       & 50 \\
ART-Type & 33.33 & 27.78 & 50  & 33.33  & 16.67    & 50 \\
Unknown &  33.33 & 27.78 & 25  & 33.33  & 33.33    & 0  \\
\bottomrule
\label{tab:clusterMov}
\end{tabular}
\end{table}

\section{Discussion}
We have collected multi-modal data with the SDS to extract digital biomarkers. With these, we aimed to distinguish PD patients from other movement disorders and find subgroups within the PD patients. We therefore evaluated two ML tasks: classification and clustering.

For the classification, we reported results based on classifiers that were trained on one data modality only. Classifiers for movement and questionnaire data generally performed better than those utilizing voice or finger tapping data. One argument for this observation is that significantly more training samples were available for questionnaire and movement data (see \autoref{tab:0}). When using a single modality, the questionnaire classifier achieved the highest balanced accuracy in both classification task with 84.3\% for PD vs. HC and 66.7\% for PD vs. DD. Combining the data modalities improved performance, resulting in a balanced classification accuracy of 91.8\% for PD vs. HC and 72.2\% for PD vs. DD. These results support our hypothesis that the proposed data recordings add informational value to the system, allowing more accurate discrimination of PD from healthy controls and, in particular, from other movement disorders. Further, we observed that PD vs. DD generally yield far less accurate classification results, indicating that more research is needed to precisely characterize and distinct PD from other similar disorders.  

In clustering, we analyzed the optimal cluster number and cluster composition for single- and multi-modal data. Single-modal data resulted in an ideal cluster number of two. Cluster S2 contained mostly the AR-type, as well as the ART-type and samples labeled Unknown. It did not, however, contain the T-type. In contrast, Cluster S1 did not contain the AR-type, it did, however, include the T- and ART-type, as well as samples labeled Unknown. The results indicate that smartwatch based features capture the clinically established PD types. The cluster analysis with multiple data modalities resulted in a finer subdivision of participants. Cluster M2 included the same samples as cluster S1. The remaining samples formed three additional cluster. Cluster M4 consisted equally of the AR- and the ART-type, whereas cluster M1 and cluster M3 consisted of the AR-type, the ART-type and samples labeled Unknown. In cluster M1, the ART-type prevailed, while cluster M3 mostly consisted of the AR-type.
The cluster analysis showed that an increase from single to multiple data modalities results in an increase in the number of clusters. Because each cluster grouped at least two different PD types, we hypothesize that clusters cannot be explained by the clinically established PD types alone. 

A limitation of our analysis is the relatively small sample size for the clustering within PD patients. Therefore, to find a stable and representative set of digital biomarkers, further evaluation with more multi-modal measurements - preferably > 200 PD participants and as many controls - is required.

\section{Conclusion}
We have conducted a study with a multi-modal recording system based on mobile smart devices to research a broad phenotypical spectrum of PD. In this work, we evaluated the information gain that results from using data from different modalities, including questionnaires, movement recordings, voice captures and smart-phone based finger tapping. Our ML analysis resulted in two main findings. 

First, combining information from different sensor sources of smart devices improved classification accuracy when distinguishing PD from the HC group. More importantly, we have seen a similar improvement in the classification between PD and the DD group, which consists of other movement disorders. These results indicate that the different data modalities complement each other and in this way aid in characterizing PD more precisely when comparing it to other disorders.

The second observation is related to the cluster analysis. Our methods have shown that we were able to identify certain subgroups within the PD group when utilizing movement data. These representations are in line with medical expectation as PD is medically categorized based on movement symptoms. However, when adding additional data modalities to the clustering, we observed a finer subdivision between clusters. This observation indicates that there are potentially more PD sub-phenotypes beyond the well-known movement-based classifications.

Finding and specifying such yet unknown groups could strongly aid in more personalized PD treatment. As our system is fully based on consumer-grade devices, it could easily be integrated to support early diagnosis and disease monitoring by giving relevant indications from combinatory digital biomarkers.

\bibliographystyle{unsrt}  
\bibliography{references}

\begin{thebibliography}{10}

\bibitem{postuma2019prodromal}
Ronald~B Postuma.
\newblock Prodromal parkinson disease: do we miss the signs?
\newblock {\em Nature Reviews Neurology}, 15(8):437--438, 2019.

\bibitem{lee2016validation}
Chae~Young Lee, Seong~Jun Kang, Sang-Kyoon Hong, Hyeo-Il Ma, Unjoo Lee, and
  Yun~Joong Kim.
\newblock A validation study of a smartphone-based finger tapping application
  for quantitative assessment of bradykinesia in parkinson’s disease.
\newblock {\em PloS one}, 11(7):e0158852, 2016.

\bibitem{rusz2018smartphone}
Jan Rusz, Jan Hlavni\v{c}ka, Tereza Tykalov\'{a}, Michal Novotn\'y, Petr
  Du\v{s}ek, Karel \v{S}onka, and Ev\v{z}en R\r{u}\v{z}i\v{c}ka.
\newblock Smartphone allows capture of speech abnormalities associated with
  high risk of developing parkinson’s disease.
\newblock {\em IEEE transactions on neural systems and rehabilitation
  engineering}, 26(8):1495--1507, 2018.

\bibitem{carignan2015measuring}
Benoit Carignan, Jean-Fran{\c{c}}ois Daneault, and Christian Duval.
\newblock Measuring tremor with a smartphone.
\newblock In {\em Mobile Health Technologies}, pages 359--374. Springer, 2015.

\bibitem{varghese2021sensor}
Julian Varghese, Catharina Marie~van Alen, Michael Fujarski, Georg~Stefan
  Schlake, Julitta Sucker, Tobias Warnecke, and Christine Thomas.
\newblock Sensor validation and diagnostic potential of smartwatches in
  movement disorders.
\newblock {\em Sensors}, 21(9):3139, 2021.

\bibitem{bhidayasiri2012parkinson}
Roongroj Bhidayasiri and Daniel Tarsy.
\newblock Parkinson’s disease: Hoehn and yahr scale.
\newblock In {\em Movement Disorders: A Video Atlas}, pages 4--5. Springer,
  2012.

\bibitem{chaudhuri2006international}
Kallol~Ray Chaudhuri, Pablo Martinez-Martin, Anthony~HV Schapira, Fabrizio
  Stocchi, Kapil Sethi, Per Odin, Richard~G Brown, William Koller, Paolo
  Barone, Graeme MacPhee, et~al.
\newblock International multicenter pilot study of the first comprehensive
  self-completed nonmotor symptoms questionnaire for parkinson's disease: the
  nmsquest study.
\newblock {\em Movement disorders: official journal of the Movement Disorder
  Society}, 21(7):916--923, 2006.

\bibitem{varghese2019smart}
Julian Varghese, Stephan Niew{\"o}hner, I{\~n}aki Soto-Rey, Stephanie
  Schipmann-Mileti{\'c}, Nils Warneke, Tobias Warnecke, and Martin Dugas.
\newblock A smart device system to identify new phenotypical characteristics in
  movement disorders.
\newblock {\em Frontiers in neurology}, 10:48, 2019.

\bibitem{sklearn_api}
Lars Buitinck, Gilles Louppe, Mathieu Blondel, Fabian Pedregosa, Andreas
  Mueller, Olivier Grisel, Vlad Niculae, Peter Prettenhofer, Alexandre
  Gramfort, Jaques Grobler, Robert Layton, Jake VanderPlas, Arnaud Joly, Brian
  Holt, and Ga{\"{e}}l Varoquaux.
\newblock {API} design for machine learning software: experiences from the
  scikit-learn project.
\newblock In {\em ECML PKDD Workshop: Languages for Data Mining and Machine
  Learning}, pages 108--122, 2013.

\bibitem{dorogush2018catboost}
Anna~Veronika Dorogush, Vasily Ershov, and Andrey Gulin.
\newblock Catboost: gradient boosting with categorical features support.
\newblock {\em arXiv preprint arXiv:1810.11363}, 2018.

\end{thebibliography}

\end{document}